\documentclass[12pt,reqno]{amsart}
\usepackage{amsaddr}

\usepackage[top=30truemm,bottom=30truemm,left=25truemm,right=25truemm]{geometry}

\usepackage{amsmath,amssymb,amsthm,mathabx}
\usepackage{bm}
\usepackage{graphicx,here,comment}
\usepackage{microtype}
\usepackage{subfigure}
\usepackage{booktabs} 
\usepackage{algorithm}
\usepackage{algorithmic}
\usepackage{multirow}
\usepackage{booktabs} 
\usepackage[stable]{footmisc}

\usepackage{amsmath,amssymb,amsthm,mathabx,amsfonts}
\usepackage{nicefrac}       
\usepackage{bm}
\usepackage{graphicx,here,comment,float,wrapfig}
\usepackage{microtype}
\usepackage{subfigure}
\usepackage{booktabs} 

\theoremstyle{definition}

\usepackage{color}

\title[Double Kernelized Scoring and Matrix Kernels]{Anomaly Detection and Localization \\ based on Double Kernelized Scoring and Matrix Kernels}
\author{Shunsuke Hirose, Tomotake Kozu \and Yingzi Jin}
\address{Deloitte Touche Tohmatsu LLC\footnote{
A translation of ``Anomaly Detection based on Doubly-Kernelized Scoring and Matrix Kernels'' by Shunsuke Hirose and Tomotake Kozu~(This article is written in Japanese). DOI:  https://doi.org/10.1527/tjsai.AI30-D}}
\email{shunsuke.hirose@tohmatsu.co.jp}

\begin{document}

\maketitle

\begin{abstract}
Anomaly detection is necessary for proper and safe operation of large-scale systems consisting of multiple devices, networks, and/or plants. 
Those systems are often characterized by a pair of multivariate datasets. 
To detect anomaly in such a system and localize element(s) associated with anomaly, one would need to estimate scores that quantify anomalousness of the entire system as well as its elements. 
However, it is not trivial to estimate such scores by considering changes of relationships between the elements, which strongly correlate with each other. 
Moreover, it is necessary to estimate the scores for the entire system and its elements from a single framework, in order to identify relationships among the scores for localizing elements associated with anomaly. 
Here, we developed a new method to quantify anomalousness of an entire system and its elements simultaneously.
\par
The purpose of this paper is threefold. 
The first one is to propose a new anomaly detection method: Double Kernelized Scoring~(DKS). 
DKS is a unified framework for entire-system anomaly scoring and element-wise anomaly scoring. 
Therefore, DKS allows for conducting simultaneously 1)~anomaly detection for the entire system and 2)~localization for identifying faulty elements responsible for the system anomaly. 
The second purpose is to propose a new kernel function: Matrix Kernel. 
The Matrix Kernel is defined between general matrices, which might have different dimensions, allowing for conducting anomaly detection on systems where the number of elements change over time. 
The third purpose is to demonstrate the effectiveness of the proposed method experimentally. 
We evaluated the proposed method with synthetic and real time series data. 
The results demonstrate that DKS is able to detect anomaly and localize the elements associated with it successfully.
\end{abstract}

%
\section{Introduction \label{introduction}}
\par
Stable operation in large-scale systems~(e.g. factory operation) is necessary to maintain safe environment, because anomalies in such a system could result in severe losses. 
In case that anomaly or failure in the system emerged, it would be desirable to predict or detect anomalies as quickly as possible. 
To suppress such losses, it would also be necessary to localize elements associated with anomaly to fix and control the system properly. 
\par
Here, we consider performing anomaly detection and localization in an unsupervised fashion from multivariate datasets where each element correlates with each other.
{\it Anomaly detection} is to decide on whether the entire system~(i.e. the entire elements included in the multivariate dataset) is anomalous or not. 
{\it Localization} is to identify faulty elements which are responsible 
for the system anomaly.
In order to solve these tasks simultaneously, it is necessary to conduct anomaly detection and localization in a unified framework. 
If anomaly detection and localization were conducted separately, it would be difficult to conduct localization as it would make the relation between the system-level anomaly and the element-wise anomaly unclear.
%
\par
Here we also consider the case in which the number of elements change over time. 
The scope of the application would be limited if such a case was not considered because it is not unusual that the number of elements fluctuates in anomaly detection. 
For example, we examine the case of anomaly detection from a communications server network. 
The system is the entire network. Each element is a communications server. 
Suppose that we want to determine whether the servers in a certain area are anomalous or not. 
In such cases, the number of elements in the entire system could be altered because of servers newly added or deleted, resulting from malfunction, for example.
\par
The purpose of this paper is threefold. 
The first one is to propose a new anomaly detection method: Double Kernelized Scoring~(DKS).
DKS is a unified method for performing anomaly detection and localization simultaneously in a strongly correlating system. 
To the best of our knowledge, this paper is the first to describe a method for detecting simultaneously entire system anomalies and elements responsible for them using a single framework. 
The key ideas of DKS are the following. 
First, we present a dataset using a kernel matrix, an element of which represents a relation between a pair of variables. 
Second, by combining the kernel between variables and a kernel between matrices, we construct a statistically natural measure that represents degree of change in the relation of variables, which corresponds to the anomalousness. 
The measure is definable between two variable groups with any number of variables.
Thus, it allows us to conduct simultaneously anomaly detection by estimating the measure between a pair of datasets and localization by estimating the measure between a pair of variables.
\par
The second one is to propose a new kernel function, which we named Matrix Kernel. 
The Matrix Kernel is defined using two matrices to estimate the anomalousness in DKS. 
In order to make DKS applicable to the aforementioned problem, we construct a kernel so that it has the following properties. 
First, the input matrices are general matrices, which are not restricted to those representing weighted graphs~(ones with non-negative matrix elements). 
Second, the inputs may have different dimensions. 
Therefore, by using the Matrix Kernel, DKS is applicable to systems where the number of elements may change over time. 
Third, the Matrix Kernel is invariant under permutation of the input matrix element index. 
Therefore, the kernel is insensitive to non-essential changes such as the permutation.
\par
The third one is to demonstrate the effectiveness of 
the proposed method by the experimental results 
using three datasets. 
Past studies have been conducted to 
analyze univariate time series 
for conducting anomaly detection~\cite{YT02,SFT17}.
However, real-world systems often consist of 
multivariate time series that have strong mutual correlation. 
In such cases, it would be difficult to detect anomalies if each of time series was monitored separately.
Therefore, we applied our method to multivariate datasets for performing anomaly detection and localization simultaneously.
\subsection{Related work}
\par
An element-wise anomaly scoring method has been proposed by using sparse structure learning of covariance matrices, 
considering correlations between time series~\cite{ILA09}.
This method detects changes of conditional probabilities of univariate time series, given the other time series as anomalies.
Another anomaly detection method was proposed for a pair of elements~\cite{MZR15}. 
This method detects the change of the relationship between a pair of univariate time series as anomalies.
Anomaly detection methods for an entire system have also been proposed~\cite{IK04,HYN09}.
These methods detect outliers of eigenvectors or eigenvalues 
as anomalies. 
%
\par
These previously proposed methods have two critical restrictions. 
First, they cannot conduct anomaly detection of the entire system and anomaly localization (i.e. anomaly detection of a single element) simultaneously because the methods detect one of the followings: anomaly of the entire system, anomaly of an element, and anomaly of a subset of elements. 
Second, the previous methods are restricted to treating fixed dimensions.
\par
For treating inputs with different dimensions, 
graph-based anomaly detection methods have been proposed~\cite{HMT15,MMA16}.
These methods represent input datasets as graphs 
and conduct anomaly detection by comparison of the graphs.
It becomes possible to treat datasets for which the numbers of elements change by using the graphs.
However, this method is unable to conduct anomaly detection and localization simultaneously from a single framework.
In the case described in Ref.~~\cite{HMT15}, 
an anomalous subgraph is detected. 
Therefore, anomalies of the entire system cannot be detected. 
In the case of Ref.~\cite{MMA16}, 
anomalies of the entire system are detected.
For that reason, localization cannot be conducted. 
\par
Kernel functions also have often been used to handle inputs with different dimensions.
Previous studies proposed kernel functions defined between weighted graphs~\cite{Hau99,KTI03}. 
They are constructed based on 
counting subtrees or paths.
Therefore, they are applicable to graphs having different numbers of nodes. 
However, it is difficult to apply them to general matrices.
In Ref.~\cite{KSK09}, 
kernels defined between matrices 
were constructed 
based on a group theoretical approach. 
To estimate the kernel value, {\it phantom nodes} were introduced. 
They are virtual and are introduced to 
transform the input matrices to constant dimensional square matrices.
Consequently, it becomes difficult to handle general matrices 
when the upper limit of the input matrix dimension~(the {\it constant} dimension) is not given in advance.
In Ref.~\cite{JKH04}, 
a kernel between general matrices has been proposed,
the Probability Product Kernel~(PPK), which is defined as an inner product 
between the probability density functions of matrix elements. 
By definition, PPK loses the information of the matrix structure.
\par
The remainder of this paper is organized as follows.
Section~\ref{proposed} 
provides a problem setting and 
proposes an anomaly detection and localization method. 
Section~\ref{mkernel} 
proposes the Matrix Kernel. 
Section~\ref{experiment} 
explains the experimentally obtained results.
Section~\ref{summary} 
presents concluding remarks.

\section{Anomaly Detection based on Double Kernelized Scoring}\label{proposed}
\par
In this section, 
we propose a method for solving the task of 
simultaneously performing anomaly detection and localization 
in multivariate time series in which the 
number of elements might change over time.
\subsection{Problem Setting}\label{problem}
Suppose that 
two multivariate datasets are given as 
\begin{eqnarray}
&&
{\mathcal D}\,=\,\left\{z_{1},z_{2},{\cdots},z_{d}\right\}, 
\,\,\,\,
{\mathcal D}'\,=\,\left\{z'_{1},z'_{2},{\cdots},z'_{d'}\right\}, 
\label{eqd02}
\end{eqnarray}
where ${\mathcal D}$ and $z_{i}$  
represent 
a dataset and a variable respectively.
The numbers of variables, $d$ and $d'$, may be different.
Hereinafter, we denote a set of all variables as {\it system}.
\par
Assume that a set of {\it target} variables is given for each dataset. 
In this paper, {\it target} does not represent a dependent variable 
used in supervised learning, 
but represents variables for which we want to measure anomalousness 
as discussed below.
We denote all the variables as follows; 
\begin{eqnarray}
&&
\boldsymbol{z}
=
(\boldsymbol{z}_{t},\boldsymbol{z}_{\bar{t}})^{T},
\,\,\,\,
\boldsymbol{z}'
=
(\boldsymbol{z}'_{t'},\boldsymbol{z}'_{\bar{t'}})^{T}, 
\end{eqnarray}
where $\boldsymbol{z}$ and $\boldsymbol{z}'$ 
represent 
a set of all variables in ${\mathcal D}$ and 
that in ${\mathcal D}'$ respectively. 
$\boldsymbol{z}$~($\boldsymbol{z}'$) is divided into 
$\boldsymbol{z}_{t}$ and $\boldsymbol{z}_{\bar{t}}$
~($\boldsymbol{z}'_{t'}$ and $\boldsymbol{z}'_{\bar{t'}}$), 
where 
$t$ and $\bar{t}$~($t'$ and $\bar{t'}$) 
represent a set of target variables 
in dataset ${\mathcal D}$~(${\mathcal D}'$) 
and its complement respectively.
\par
For a given pair of datasets and given sets of target variables, 
we want to solve the problem of 
performing anomaly detection and localization simultaneously.
We define the problem as one consisting of 
the following two tasks. 
The first one is 
to estimate a system anomaly score, $S(\boldsymbol{z},\boldsymbol{z}')$ 
that represents the degree to which the system is anomalous~(i.e. the higher anomaly score indicates that the system is more anomalous).

The second one is 
to estimate a set of {\it target anomaly scores}, 
$\{S_{tt'}(\boldsymbol{z},\boldsymbol{z}')\}_{t,t'}$, 
which represents how anomalous variable set $(t,t')$ is.
The higher target anomaly score of the variable set $(t,t')$ 
indicates that the set is more likely to be responsible for the system anomaly.
For example, 
if $t=t'$ and $t$ represents a single variable $z$, 
then $S_{tt'}(\boldsymbol{z},\boldsymbol{z}')$ represents 
how anomalous variable $z$ is.
\par
We propose that the anomaly scores should satisfy the following requirements. 
First, the system anomaly score $S(\boldsymbol{z},\boldsymbol{z}')$ 
and target anomaly score $S_{tt'}(\boldsymbol{z},\boldsymbol{z}')$ 
are estimated using a single framework. 
If they are estimated from different ones, 
it becomes difficult to conduct localization 
because the relation between the scores is unclear.
Second, 
the anomaly scores are statistically natural measures 
representing difference between the target sets, $t$ and $t'$.
\subsection{Double Kernelized Scoring}
In this section, we propose a method for solving the aforementioned problem.
The method estimates the anomaly scores 
using kernel functions of two kinds. 
Therefore we designate the method as {\it Double Kernelized Scoring}~(DKS).
The overall flow of DKS is summarized as follows. 
\par
First, input is a pair of multivariate datasets, 
${\mathcal D}$ and ${\mathcal D}'$. 
Here ${\mathcal D}$ and ${\mathcal D}'$ consist of 
$d$ and $d'$ variables respectively as in Section~\ref{problem}.
\par
From the datasets, 
we derive a pair of kernel matrices, 
whose element is defined between variables as follows:
\begin{eqnarray}
&&
{\mathcal D}
\,{\rightarrow}\,
\left\{
K(z_{i},z_{j})
\right\}_{i,j=1}^{d}, 
\,\,\,\,
{\mathcal D}'
\,{\rightarrow}\,
\left\{
K'(z'_{i},z'_{j})
\right\}_{i,j=1}^{d'}.
\end{eqnarray}
Note that the kernel is defined between {\it variables}, not between observations.
For example, 
we can use 
a covariance matrix and a correlation coefficient matrix 
as a kernel matrix.
\par
We now define scoring targets $\{(t, t')\}$
~(see Section~\ref{problem}) consisting of some variables 
for which we want to estimate anomaly score.
For estimating variable-wise anomaly scores, 
we define the targets as 
$\{(t, t')\}=\{(z_{1},z'_{1}),{\cdots},(z_{d},z'_{d})\}$, 
where $t$ and $t'$ represent the same single variable.
On the other hand,for estimating a system anomaly score, 
we define the target as 
$\{(t, t')\}=(\boldsymbol{z},\boldsymbol{z}')$, 
where $t$ and $t'$ 
represent a set of all variables in ${\mathcal D}$ and ${\mathcal D}'$ respectively.
The numbers of variables in $t$ and $t'$ may be different.
Corresponding to $(t,t')$, 
we designate the kernel matrices as
\begin{eqnarray}
&&
K
\,=\,
\left(
\begin{array}{cc}
K_{tt}&K_{t\bar{t}}\\
K_{\bar{t}t}&K_{\bar{t}\bar{t}}
\end{array}
\right),
\,\,\,\,
K'
\,=\,
\left(
\begin{array}{cc}
K'_{t't'}&K'_{t'\bar{t'}}\\
K'_{\bar{t'}t'}&K'_{\bar{t'}\bar{t'}}
\label{divker02}
\end{array}
\right), 
\end{eqnarray}
where $\bar{t}$ and $\bar{t'}$ represent 
a complement of $t$ and that of $t'$ respectively.
\par
Next, we estimate the anomaly score 
which corresponds to $t$ and $t'$ as 
\begin{eqnarray}
S_{tt'}(\boldsymbol{z},\boldsymbol{z}')
&=&
\bigl[
K_{M}(K,K'^{-1})
+K_{M}(K',K^{-1})
-K_{M}(K,K^{-1})
-K_{M}(K',K'^{-1})
\bigr]
\nonumber
\\
&&
-
\bigl[
K_{M}(K_{\bar{t}\bar{t}},K_{\bar{t'}\bar{t'}}^{'-1})
+K_{M}(K'_{\bar{t'}\bar{t'}},K_{\bar{t}\bar{t}}^{-1})
-K_{M}(K_{\bar{t}\bar{t}},K_{\bar{t}\bar{t}}^{-1})
-K_{M}(K'_{\bar{t'}\bar{t'}},K_{\bar{t'}\bar{t'}}^{'-1})
\bigr], 
\hspace{2mm}
\label{eqscr08}
\end{eqnarray}
where $K_{M}$ represents a kernel defined between matrices.
As described in Section~\ref{mkernel}, we introduce the Matrix Kernel that is defined between matrices with different dimensions. 
Using the Matrix Kernel for $K_{M}$, 
the anomaly score can be estimated 
even when $t$ and $t'$ have different dimensions. 
We derive the anomaly scores for this case in 
Section~\ref{secderivation}.
\par
Finally, we conduct anomaly detection and localization 
by using the anomaly scores.  
If the system anomaly score is high, 
then we regard the system as anomalous.
If the system anomaly score and 
target anomaly scores are high, 
then the variables in the targets 
are regarded as being responsible for 
system anomalies.

\subsection{Derivation of Anomaly Scores}\label{secderivation}
\par
In this section, 
we derive the anomaly score defined in Eq.~(\ref{eqscr08}). 
\par
We define a system anomaly score 
as a distance between kernel matrices, $D(K,K')$ as follows: 
\begin{eqnarray}
S(\boldsymbol{z},\boldsymbol{z}')
&=&
D(K,K').
\label{eqscr00}
\end{eqnarray}
The distance between kernels 
represents the amount of change 
of relations between variables. 
Therefore this score is a natural measure 
that represents how anomalous the system is.
Next, we generalize the score in Eq.~(\ref{eqscr00}) 
and define a target score as 
\begin{eqnarray}
S_{tt'}(\boldsymbol{z},\boldsymbol{z}')
&=&
D(K,K')-D(K_{\bar{t}\bar{t}},K'_{\bar{t'}\bar{t'}}).
\label{eqscr01}
\end{eqnarray}
The system anomaly score in Eq.~(\ref{eqscr00}) 
is included in the definition of a target anomaly score in Eq.~(\ref{eqscr01}) 
because 
$
S(\boldsymbol{z},\boldsymbol{z}')
\,=\,
S_{tt'}(\boldsymbol{z},\boldsymbol{z}')\bigl|_{\bar{t}=\bar{t'}={\phi}}
\,=\,D(K,K')
$
holds. 
Here, ${\phi}$ represents an empty set.
\par
Using the scores in 
Eq.~(\ref{eqscr01}) and Eq.~(\ref{eqscr00}), 
we can estimate both a system anomaly score and 
target anomaly scores that includes variable-wise anomaly scores 
from a single framework.
That is, 
we can estimate an anomaly score for 
a set with any number of variables, 
from a single variable to all variables, 
by using the score in Eq.~(\ref{eqscr01}).
\par
As a distance function defined between kernels, 
we use the symmetrized Burg divergence~\cite{KSD06}:
\begin{eqnarray}
D(K,K')
&=&
D_{\mbox{B}}(K||K')+D_{\mbox{B}}(K'||K), 
\\
D_{\mbox{B}}(X||Y)
&=&
\mbox{tr}[XY^{-1}]-{\log}|XY^{-1}|+m, 
\end{eqnarray}
where $D_{B}$ and $m$ represent 
Burg divergence and the dimension of matrix $X$ respectively.
Then the target anomaly score becomes
\begin{eqnarray}
S_{tt'}(\boldsymbol{z},\boldsymbol{z}')
&=&
\mbox{tr}[KK'^{-1}]
+\mbox{tr}[K'K^{-1}]
-\mbox{tr}[KK^{-1}]
-\mbox{tr}[K'K'^{-1}]
\nonumber \\
&&
\hspace{5mm}
-\mbox{tr}[K_{\bar{t}\bar{t}}K_{\bar{t'}\bar{t'}}^{'-1}]
-\mbox{tr}[K'_{\bar{t'}\bar{t'}}K_{\bar{t}\bar{t}}^{-1}]
+\mbox{tr}[K_{\bar{t}\bar{t}}K_{\bar{t}\bar{t}}^{-1}]
+\mbox{tr}[K'_{\bar{t'}\bar{t'}}K_{\bar{t'}\bar{t'}}^{'-1}]
\label{eqscr03}
\end{eqnarray}
For deriving Eq.~(\ref{eqscr03}), 
we used $m=\mbox{tr}[XX^{-1}]$ 
and ${\log}|XY^{-1}|+{\log}|YX^{-1}|=0$.
$K_{\bar{t}\bar{t}}^{-1}$ represents 
an inverse of submatrix $K_{\bar{t}\bar{t}}$.
A system anomaly score is derived 
using 
$
S(\boldsymbol{z},\boldsymbol{z}')
\,=\,
S_{tt'}(\boldsymbol{z},\boldsymbol{z}')\bigl|_{\bar{t}=\bar{t'}={\phi}}
\,=\,D(K,K'). 
$
\par
We show that 
the anomaly score in Eq.~(\ref{eqscr03}) 
is a natural measure 
that represents the change amount between $t$ and $t'$. 
We denote 
a feature vector in a kernel space as ${\psi}$ 
and denote its ${\alpha}$-th component as ${\psi}_{\alpha}$: 
$
K(z_{i},z_{j})
=
{\sum_{\alpha}}{\psi}_{\alpha}(z_{i}){\psi}_{\alpha}(z_{j}).
$
By using a vector $\boldsymbol{w}$ 
that follows a standard normal distribution,
we construct variables 
$\{y_{i}\}$ and $\{y'_{j}\}$ 
from $\{z_{i}\}$ and $\{z'_{j}\}$ as
\begin{eqnarray}
&&
\boldsymbol{w}
\,{\sim}\,
{\mathcal N}(\boldsymbol{w}|\boldsymbol{\mu}={\bf 0},{\Sigma}=I), 
\label{eqvar01_00}
\\
&&
y_{i}
=
{\sum_{\alpha}}w_{\alpha}{\psi}_{\alpha}(z_{i}),\,\,\,
y'_{j}
=
{\sum_{\alpha}}w_{\alpha}{\psi}_{\alpha}(z'_{j}).
\label{eqvar01}
\end{eqnarray}
Vectors $\boldsymbol{y}$ and $\boldsymbol{y}'$, of which the $i$-th components 
are $y_{i}$ and $y'_{i}$ respectively, 
follow multivariate normal distributions as
\begin{eqnarray}
&&
\boldsymbol{y}
\,{\sim}\,
p(\boldsymbol{y})
\,=\,
{\mathcal N}(\boldsymbol{y}|{\bf 0},K), 
\,
\boldsymbol{y}'
\,{\sim}\,
p'(\boldsymbol{y}')
\,=\,
{\mathcal N}(\boldsymbol{y}'|{\bf 0},K'). 
\hspace{-5mm}
\end{eqnarray}
If one assumes that $t=t'$ and $\bar{t}=\bar{t'}$ hold, 
then the following relationship 
among the distributions of $\boldsymbol{y}$ and $\boldsymbol{y}'$, 
and the anomaly scores hold:  
\begin{eqnarray}
S_{tt'}(\boldsymbol{z},\boldsymbol{z}')
&=&
2E_{p(\boldsymbol{y}_{\bar{t}})}
\left[
D_{\mbox{KL}}\left(
p(\boldsymbol{y}_{t}|\boldsymbol{y}_{\bar{t}})
||
p'(\boldsymbol{y}_{t}|\boldsymbol{y}_{\bar{t}})
\right)
\right]
+2E_{p'(\boldsymbol{y}_{\bar{t}})}
\left[
D_{\mbox{KL}}\left(
p'(\boldsymbol{y}_{t}|\boldsymbol{y}_{\bar{t}})
||
p(\boldsymbol{y}_{t}|\boldsymbol{y}_{\bar{t}})
\right)
\right]
\nonumber
\\
&=&
2
{\int}\!\mbox{d}\boldsymbol{y}_{\bar{t}}\,
p(\boldsymbol{y}_{\bar{t}})
{\int}\!\mbox{d}\boldsymbol{y}_{{t}}\,
p(\boldsymbol{y}_{t}|\boldsymbol{y}_{\bar{t}})
{\log}\frac{
p(\boldsymbol{y}_{t}|\boldsymbol{y}_{\bar{t}})
}{
p'(\boldsymbol{y}_{t}|\boldsymbol{y}_{\bar{t}})
}
+
2
{\int}\!\mbox{d}\boldsymbol{y}_{\bar{t}}\,
p'(\boldsymbol{y}_{\bar{t}})
{\int}\!\mbox{d}\boldsymbol{y}_{{t}}\,
p'(\boldsymbol{y}_{t}|\boldsymbol{y}_{\bar{t}})
{\log}\frac{
p'(\boldsymbol{y}_{t}|\boldsymbol{y}_{\bar{t}})
}{
p(\boldsymbol{y}_{t}|\boldsymbol{y}_{\bar{t}})
},
\hspace{10mm} 
\label{eqscr04}
\end{eqnarray}
where $D_{KL}(p||p')$ represents a KL divergence 
between probability density functions~(pdfs) $p$ and $p'$.
Corresponding to the representation
$\boldsymbol{z}=(\boldsymbol{z}_{t},\,\boldsymbol{z}_{\bar{t}})$, 
we designated $\boldsymbol{y}$ as $\boldsymbol{y}=(\boldsymbol{y}_{t},\,\boldsymbol{y}_{\bar{t}})$.
We omit the proof of Eq.~(\ref{eqscr04}) because of space limitations, 
but it would be easy to derive it 
as the KL divergence in Eq.~(\ref{eqscr04}) 
is analytically tractable for Gaussians.
From Eq.~(\ref{eqscr04}), 
the anomaly score of $(t,t')$, Eq.~(\ref{eqscr03}),
represents 
the expected KL divergence between conditional probabilities 
$p(\boldsymbol{y}_{t}|\boldsymbol{y}_{\bar{t}})$ and $p'(\boldsymbol{y}_{t}|\boldsymbol{y}_{\bar{t}})$, 
integrated over the distributions 
$p(\boldsymbol{y}_{\bar{t}})$ or $p'(\boldsymbol{y}_{\bar{t}})$.
Thus, 
the anomaly score in Eq.~(\ref{eqscr03}) 
is a natural measure 
that represents the change amount between $t$ and $t'$
because the score has a clear interpretation that 
it represents the change amount between pdfs
in the space of $\boldsymbol{y}$.
\par
The anomaly score in Eq.~(\ref{eqscr03}) 
is definable
only when $dim(t)=dim(t')$ and $dim(\bar{t})=dim(\bar{t'})$ hold, 
where $dim(t)$ represents a number of variables in a variable set $t$.
Below, we eliminate this limitation and generalize the score.
The anomaly score is represented as 
a sum of traces of matrix products. 
A trace of a matrix product, $\mbox{tr}[X^{T}Y]$, 
is equal to an inner product of vectorized matrices:
$
\mbox{tr}[X^{T}Y]
\,=\,{\sum_{ij}}X^{T}_{ij}Y_{ji}
\,=\,\mbox{vec}(X){\cdot}\mbox{vec}(Y). 
$
A Kernel function computes a generalized inner product. 
Therefore, we generalize the score by introducing 
the following replacement, 
where $K_{M}$ is a kernel defined between matrices: 
$\mbox{tr}[XY]\,{\rightarrow}\,K_{M}(X,Y).$ 
By the replacement, 
the anomaly score defined in Eq.~(\ref{eqscr08}), 
$S_{tt'}(\boldsymbol{z},\boldsymbol{z}')$, is derived.
Using the kernel introduced in the next section for $K_{M}$,
we can define anomaly scores 
even when $t=t'$ or $\bar{t}=\bar{t'}$ does not hold.
If we use a dot product as $K_{M}$ 
\begin{eqnarray}
K_{M}(X,Y)
&=&
\mbox{vec}(X){\cdot}\mbox{vec}(Y)
\,=\,
\mbox{tr}[XY], 
\label{eqkernel02}
\end{eqnarray}
then the anomaly score in Eq.~(\ref{eqscr08}) 
becomes that in Eq.~(\ref{eqscr03}) inversely.
\section{Matrix Kernel}\label{mkernel}
\par
In this section, we propose {\it Matrix Kernel} 
as $K_{M}$ included in the anomaly score proposed in Section~\ref{proposed}
~(the anomaly score is defined as Eq.~(\ref{eqscr08})). 
The Matrix Kernel is a kernel defined between matrices.
%
\par
The reason that we introduce the Matrix Kernel is 
to make DKS robust and applicable to 
multivariate time series in which the number of elements might change over time. 
\subsection{Problem Setting}\label{cond01}
\par
We assume that two real matrices $A$ and $A'$ are given 
and that they are the
$d{\times}d$ matrix and $d'{\times}d'$ matrix respectively.
We aim to derive 
a kernel $K_{M}$ for which inputs are these matrices.
Under the problem setting in Section~\ref{proposed}, 
the inputs are restricted to kernel matrices. 
Therefore, we can consider $A$ and $A'$ as positive semidefinite.
\par
We require that 
the Matrix Kernel satisfies the following two conditions. 
The first condition is that input matrices may have different dimensions
~(i.e. $d{\ne}d'$ is permitted).
This condition is necessary 
to consider fluctuation of the number of elements~(i.e. the number of variables) as mentioned in Section~\ref{introduction}. 
In this case, 
either $K$ and $K'$ have different dimensions, 
or $K_{\bar{t}\bar{t}}$ and $K'_{\bar{t'}\bar{t'}}$ have different dimensions.
Therefore, 
to estimate the anomaly scores in Eq.~(\ref{eqscr08}), 
it is necessary for $K_{M}$ to satisfy the first condition.
\par
The second condition is that the Matrix Kernel has {\it permutation invariance}. 
Permutation invariance means that 
the output of the kernel does not change 
if we permute the index of the matrix elements of $A$ and $A'$ separately. 
Assume that $A$ and $A'$ are respectively transformed as 
$A_{p}$ and $A'_{p'}$ by matrix element index permutation 
\footnote{
For example, $A$ is transformed as 
\begin{eqnarray}
&&
A=
\left(
\begin{tabular}{cc}
$A_{11}$ & $A_{12}$ \\
$A_{21}$ & $A_{22}$ 
\end{tabular}
\right)
\,{\rightarrow}\,
A_{p}=
\left(
\begin{tabular}{cc}
$A_{22}$ & $A_{21}$ \\
$A_{12}$ & $A_{11}$  
\end{tabular}
\right), 
\nonumber
\end{eqnarray}
where $p$ represents a permutation of elements ``1'' and ``2''. 
}
The condition requires $K_{M}(A,A')=K_{M}(A_{p},A'_{p'})$ holds
for any pair of such permutations. 
Here $p$ and $p'$ may be different.
The second condition is necessary 
to make DKS robust.
If $K_{M}$ in Eq.~(\ref{eqscr08}) satisfies this condition, whereas 
the anomaly scores 
are invariant under non-essential changes because of the permutation
~(i.e. non-topological changes of kernel matrices because 
they can be transformed by the permutation), 
they are sensitive to topological changes of kernel matrices). 
\par
Hereinafter we designate 
these two conditions as matrix kernel conditions.
\subsection{Representation of Matrix Kernel}
\par
Suppose that the input matrices are decomposed as 
\begin{eqnarray}
&&
A
=
{\sum_{k=1}^{d}}
\boldsymbol{u}_{k}{\lambda}_{k}\boldsymbol{u}_{k}^{T},\,\,\,
A'
=
{\sum_{l=1}^{d'}}
\boldsymbol{u}_{l}'{\lambda}_{l}'\boldsymbol{u}_{l}^{`T}, 
\label{inpmat02}
\end{eqnarray}
where 
${\lambda}_{k}$ and $\boldsymbol{u}_{k}$~(${\lambda}'_{k}$ and $\boldsymbol{u}'_{k}$) represent the $k$-th eigenvalue and the $k$-th eigenvector of the input matrix $A$~($A'$) respectively.
We assume that 
the eigenvalue decompositions in Eq.~(\ref{inpmat02}) 
become unique 
if we require additional conditions such as 
``$\boldsymbol{u}_{k}^{T}\boldsymbol{u}_{k}=1$'', 
``${\sum_{i=1}^{d}}(\boldsymbol{u}_{k})_{i}{\ge}0$'', 
and 
``If ${\lambda}_{k}={\lambda}_{k+1}$, 
then we replace $\boldsymbol{u}_{k}$ with $\boldsymbol{v}_{k}=c_{k}\boldsymbol{u}_{k}+c_{k+1}\boldsymbol{u}_{k+1}$, where $\boldsymbol{v}_{k}$ maximizes ${\sum_{i=1}^{d}}|(\boldsymbol{v}_{k})_{i}|$''.
Here $(\boldsymbol{u}_{k})_{i}$ represents the $i$-th component of $\boldsymbol{u}_{k}$.
\par
We define the Matrix Kernel by Eq.~(\ref{mkdir02}) below. 
For experimental results in Section~\ref{experiment}, we used the following Matrix Kernel derived for the univariate normal distribution:  
\begin{eqnarray}
K_{M}(A,\,A')
&=&
{\sum_{k=1}^{d}}
{\sum_{l=1}^{d'}}
{\lambda}_{k}{\lambda}_{l}'
\left(
\frac{
2{\sigma}(\boldsymbol{u}_{k}){\sigma}(\boldsymbol{u}_{l}')
}{
{\sigma}(\boldsymbol{u}_{k})^{2}+{\sigma}(\boldsymbol{u}_{l}')^{2}
}
\right)
{\exp}\left[
-\frac{
(
{\mu}(\boldsymbol{u}_{k})-{\mu}(\boldsymbol{u}_{l}')
)^{2}
}{
2(
{\sigma}(\boldsymbol{u}_{k})^{2}+{\sigma}(\boldsymbol{u}_{l}')^{2}
)
}
\right], 
\label{matker02}
\end{eqnarray}
where 
${\sigma}(\boldsymbol{x})$ and ${\mu}(\boldsymbol{x})$ represent 
the standard deviation of vector $\boldsymbol{x}$'s components
and the mean of the components respectively.

\subsection{Derivation of Matrix Kernel}\label{mkderivation}
In this section, we derive the kernel in Eq.~(\ref{matker02}).
\par
If the dimensions of $A$ and $A'$ ($d$ and $d'$ in Eq.~(\ref{inpmat02})) 
are the same, then
we can define the following kernel function 
represented as an inner product:
\begin{eqnarray}
&&
K_{I}(A,\,A')
=
{\sum_{i,j=1}^{d}}
A_{ij}A'_{ij}
=
{\sum_{k,l=1}^{d}}
\left[{\lambda}_{k}{\lambda}_{l}'\right]
\left[\boldsymbol{u}_{k}^{T}\boldsymbol{u}_{l}'\right]^{2}, 
\hspace{-5mm}
\label{mkdir01}
\end{eqnarray}
where $(\boldsymbol{u}_{k})_{i}$ represents the $i$-th component of $\boldsymbol{u}_{k}$.
$K_{I}$ in Eq.~(\ref{mkdir01}) is a kernel function 
defined between matrices.
However, $K_{I}$ does not satisfy the matrix kernel conditions 
described in Section~\ref{cond01} 
because $K_{I}$ is not permutation invariant 
and is definable only when $d=d'$ holds.
\par
By generalizing the inner product in Eq.~(\ref{mkdir01}), 
we define a Matrix Kernel as 
\begin{eqnarray}
&&
K_{M}(A,\,A')
=
{\sum_{k=1}^{d}}{\sum_{l=1}^{d'}}
K_{s}({\lambda}_{k},{\lambda}_{l}')
\left[K_{v}(\boldsymbol{u}_{k},\boldsymbol{u}_{l}')\right]^{2}.
\label{mkdir02}
\end{eqnarray}
Eq.~(\ref{mkdir02}) is derived from Eq.~(\ref{mkdir01}) 
by replacing a scalar product with a kernel $K_{s}$ defined between scalars and 
a vector inner product with a kernel $K_{v}$ defined between vectors.
Note that 
$K_{M}$ in Eq.~(\ref{mkdir02}) is a kernel defined between matrices 
because (1)~its inputs are matrices 
and (2)~it is represented as a sum of 
kernel products with positive coefficients ($=1$). 
$K_{M}$ satisfies the matrix kernel conditions 
if both $K_{s}$ and $K_{v}$ satisfy the conditions. 
\par
As a kernel $K_{s}$, 
we use the following representation:
\begin{eqnarray}
K_{s}({\lambda}_{k},\,{\lambda}'_{l})
&=&
{\lambda}_{k}{\lambda}'_{l}. 
\label{mkdir03}
\end{eqnarray}
$K_{s}$ in Eq.~(\ref{mkdir03}) satisfies the matrix kernel conditions 
for the following reasons. 
First, ${\lambda}_{k}{\lambda}'_{l}$ can be defined independently of the range of indices $k$ and $l$.
Second, $K_{s}$ has the permutation invariance 
because eigenvalues are permutation invariant~\footnote{
The eigenvalues of matrix $A$ are invariant 
under an orthogonal transformation $U$, as we demonstrate below.  
\begin{eqnarray}
A\boldsymbol{u}_{k}
&=&
{\lambda}_{k}\boldsymbol{u}_{k}
\label{eigeq01}
\\
\left(UAU^{T}\right)\left(U\boldsymbol{u}_{k}\right)
&=&
{\lambda}_{k}\left(U\boldsymbol{u}_{k}\right)
\label{eigeq02}
\end{eqnarray}
Eq.~(\ref{eigeq02}) is derived by multiplying $U$ 
from the left of the eigenequation of $A$ 
in Eq.~(\ref{eigeq01}). Therefore, they are equal.
It is apparent that matrix $A$ and eigenvector $\boldsymbol{u}_{k}$ 
are transformed as 
$A{\rightarrow}UAU^{T}$ and $\boldsymbol{u}_{k}{\rightarrow}U\boldsymbol{u}_{k}$ 
under $U$.
However, eigenvalue ${\lambda}_{k}$ is transformed as
${\lambda}_{k}{\rightarrow}{\lambda}_{k}$, which implies that eigenvalues are invariant 
under an orthogonal transformation, 
which includes the permutation. 
Therefore eigenvalues are permutation invariant. 
}.
\par
Next, we construct $K_{v}$ such that it satisfies the matrix kernel conditions. 
We denote a pdf of $\boldsymbol{u}$'s~($\boldsymbol{u}'$'s) component  
as $p(x|\boldsymbol{u})$~($p(x|\boldsymbol{u}')$), 
where $x$ represents a scalar.
Such a pdf has the following two properties.
First, the pdf $p(x|\boldsymbol{u})$ can be regarded as 
an infinite dimensional vector 
independently of the dimension of $\boldsymbol{u}$ because 
$p(x|\boldsymbol{u})$ is a univariate function 
and because $x$ can be regarded as a vector component index.
Second, $p(x|\boldsymbol{u})$ is invariant 
under a permutation of vector component index of $\boldsymbol{u}$ because 
a pdf of $\boldsymbol{u}$'s component is independent of 
the order of the components.
\par
Using these properties, 
we construct $K_{v}$ as
\begin{eqnarray}
K_{v}(\boldsymbol{u},\boldsymbol{u}')
&=&
{\int_{-\infty}^{\infty}}\!\mbox{d}x\,
\sqrt{p(x|\boldsymbol{u})p(x|\boldsymbol{u}')}.
\label{mkdir04}
\end{eqnarray}
Eq.~(\ref{mkdir04}) is derived from 
$K_{v}=\boldsymbol{u}^{T}\boldsymbol{u}'={\sum_{i}}u_{i}u'_{i}$ 
by replacing $u_{i}$, $u'_{i}$ and ${\sum_{i}}$ 
respectively with $\sqrt{p(x|\boldsymbol{u})},\sqrt{p(x|\boldsymbol{u}')}$ 
and ${\int_{-\infty}^{\infty}}\!\mbox{d}x$.
$K_{v}$ in Eq.~(\ref{mkdir04}) has the following properties. 
First, $K_{v}$ is a kernel function 
because Eq.~(\ref{mkdir04}) is an inner product of 
infinite dimensional vectors. 
Second, $K_{v}$ satisfies the matrix kernel conditions 
because 1)~$K_{v}$ is definable between vectors even when 
$\boldsymbol{u}$ and $\boldsymbol{u}'$ have different dimensions, 
and 2)~$K_{v}$ is permutation invariant 
because $p(x|\boldsymbol{u})$ and $p(x|\boldsymbol{u}')$ are permutation invariant.
As a pdf, we use a univariate normal distribution
\footnote{
We can use any type of pdf as $p(x|\boldsymbol{u})$. 
However, for simplicity, we use a univariate normal distribution 
in this discussion.
}
:
\begin{eqnarray}
&&
p(x|\boldsymbol{u})
\,=\,
\frac{1}{\sqrt{2{\pi}}{\sigma}(\boldsymbol{u})}
{\exp}\left[
-\frac{1}{2}\left(
\frac{x-{\mu}(\boldsymbol{u})}{{\sigma}(\boldsymbol{u})}
\right)^{2}
\right], 
\label{mkdir05}
\end{eqnarray}
where ${\sigma}(\boldsymbol{u})$ and ${\mu}(\boldsymbol{u})$ represent 
the standard deviation of vector $\boldsymbol{u}$’s components 
and the mean of the components respectively.
\par
By substituting Eqs.~(\ref{mkdir03}), (\ref{mkdir04}) and (\ref{mkdir05}) 
to Eq.~(\ref{mkdir02}), 
the representation of the Matrix Kernel in Eq.~(\ref{matker02}) 
is derived.
The integration in Eq.~(\ref{mkdir04}) is analytically tractable 
because the pdfs in Eq.~(\ref{mkdir04}) are Gaussians.
Therefore Eq.~(\ref{matker02}) is derived analytically.
\section{Experiments}\label{experiment}
\par
Here, we examined the utility of DKS using three multivariate datasets.
We show that DKS can detect 
1)~element anomalies more accurately than 
an existing method~(Section~\ref{experiment01}), 
2)~changes of the numbers of elements as anomalies 
using the Matrix Kernel~(Section~\ref{experiment02}). 
and 
3)~system anomalies and elements responsible for the anomalies 
simultaneously~(Section~\ref{experiment03}). 
No standard method exists for detecting system anomalies and element anomalies simultaneously from a single framework.
Therefore, in Section~\ref{experiment03}, we did not compare DKS with any other method.

\subsection{Single Variable Anomaly Scoring}\label{experiment01}
\subsubsection{{\bf Experimental Setting}}\label{exp01}
\par
In this experiment, 
we conducted anomaly scoring for a single variable 
included in a multivariate time series. 
Then, we compared the results obtained with
DKS with those obtained from an existing method.
\par
We used {\it Synthetic Control Chart Time Series dataset}\cite{PC98,UCI}.
We used $60$ ``normal'' mode time series, 
each of which consisted of $100$ time steps, 
and 
replaced a set of their $t=51,{\cdots}100$ data points 
with that of ``cyclic'' mode time series 
randomly with probability $1/3$.
We considered 
$t=1,{\cdots},50$ data points~($60$ time series ${\times}$ $50$ time steps) 
as ${\mathcal D}$ in Eq.~(\ref{eqd02}) 
and 
$t=51,{\cdots},100$ data points 
as ${\mathcal D}'$ in Eq.~(\ref{eqd02}), respectively.
Then we estimated the anomaly scores of the individual time series.
We considered the replacement~({\it change}) as the anomaly to be detected. 
\par
As a kernel defined between variables, 
we used a covariance matrix and a diffusion kernel~\cite{KL02} 
described below.
We defined a diffusion kernel $K$ as follows: 
$
C_{ij}=(\mbox{correlation between $z_{i}$ and $z_{j}$})
$, 
$
L_{ij}=\left[{\sum_{k}}|C_{ik}|\right]$${\delta}_{ij}-|C_{ij}|
$, $K={\exp}\left[-{\lambda}L\right]$. 
Here ${\delta}_{ij}$ represents Kronecker's delta.
It becomes $1$ if $i=j$ holds, and $0$ otherwise.
Matrices $C$, $L$, and $K$ represent 
a correlation matrix, a graph Laplacian, and a diffusion kernel matrix respectively.
We fixed parameter ${\lambda}$ as ${\lambda}=1.0$.
As a kernel defined between matrices, 
we used the Matrix Kernel in Eq.~(\ref{matker02}), 
and a dot product in Eq.~(\ref{eqkernel02}).
\par
We compared the results obtained with DKS with those obtained using 
Sparse Structure Learning~(SSL)\cite{ILA09}.
SSL (1)~estimates sparse precision matrices~(inverse of covariance matrices) 
of a pair of multivariate time series, and 
(2)~estimates the anomaly scores of the individual time series 
using the sparse~(and therefore robust) structure.
SSL is not a kernel method. 
However, 
SSL estimates covariance matrices and uses a dot product~(trace) of them. 
Therefore, it can be said that 
SSL {\it corresponds} to 
a DKS where a covariance and a dot product are used  
as a kernel between variables and that between matrices respectively.
We set the free parameter of SSL as ${\rho}=0.7$, 
which represented a ratio of graphical lasso\cite{FHT08}'s 
penalty 
and 
derived the best experimental results in a study reported 
in Ref.~\cite{ILA09}.
\par
As an estimation measure, 
we used the area under the curve~(AUC) of the ROC curve.
AUC becomes high 
when the time series with the replacement have higher scores 
and those without the replacement have lower scores.
The random replacement was iterated $100$ times. 
We then evaluated the two methods, SSL and DKS, using AUC.

\subsubsection{{\bf Experimental Results}}
\begin{table}[htbp]
  \caption{Experimental results of Section~\ref{experiment01}, 
single variable anomaly scoring. 
``kernel(variables)'' and ``kernel(matrices)'' 
represent a kernel defined between variables 
and a kernel defined between matrices, which were used in the method respectively. 
``Covariance'', ``Diffusion'', ``DP'', ``PPK'', and ``MATRIX'' represent the
covariance matrix, diffusion kernel, dot product, 
probability product kernel, and the Matrix Kernel respectively. 
AUC is represented as (mean)${\pm}$(standard deviation).
The best result is presented in bold typeface.
}
\hspace{-5mm}
\begin{tabular}{|c|cc|c|}
\hline
 method & kernel(variables) & kernel(matrices) & AUC 
\\ \hline
SSL & Covariance & DP & $0.685{\pm}0.120$ 
\\ \hline
DKS & Covariance & DP & $0.659{\pm}0.113$ 
\\
DKS & Covariance & MATRIX & $0.583{\pm}0.111$ 
\\ \hline
DKS & Diffusion & DP & $0.865{\pm}0.079$ 
\\
DKS & Diffusion & MATRIX & ${\bf 0.938{\pm}0.064}$ 
\\ \hline
    \end{tabular}%
  \label{tab01}%
\end{table}%
\par
The experimentally obtained results are presented in Table~\ref{tab01}.
From the table, 
it is readily apparent that DKS is effective for 
anomaly detection for a single variable 
included in a multivariate time series for the following reasons. 
\par
First, DKS with a diffusion kernel or a Matrix Kernel 
outperformed SSL. 
It is one of DKS's own features 
that we can select kernels between variables and those between matrices,
although we cannot in the case of SSL.
Therefore it was not possible to adjust SSL to outperform DKS. 
For real-world applications, 
it is important to select optimal kernels for DKS.
Selection methods include a cross validation.
However, 
to construct a kernel selection method is 
beyond the scope of this paper.
%
\par
Second, 
DKS with a covariance matrix and a dot product was comparable to SSL.
Here, {\it comparable} means that 
their AUCs were inferred as the same 
based on the results of the $t$-test with $95\%$ confidence.
This result indicates that 
the anomaly score defined in Eq.~(\ref{eqscr08}) is 
valid for anomaly detection even in a linear space, 
which SSL considers.

\subsection{Change Detection of Constituent Variables}\label{experiment02}
\subsubsection{{\bf Experimental Setting}}\label{exp02}
\par
In this experiment, 
we used DKS for detecting anomalies where the numbers of variables change, 
to compare the Matrix Kernel with an existing kernel.
\par
Two multivariate datasets were generated randomly in which each of the variables followed a standard normal distribution and consisted of $200$ observations.
We designated the datasets as 
${\mathcal D}$ in Eq.~(\ref{eqd02})
and 
${\mathcal D}'$ in Eq.~(\ref{eqd02}). 
${\mathcal D}$ and ${\mathcal D}'$ 
consisted of $9$ and $10$ variables respectively.
\par
Under two settings presented in Table~\ref{tab03},  
we used DKS to estimate anomaly scores 
for variable groups~(``Group''s in Table~\ref{tab03}).
These settings 
had the same datasets
~(${\mathcal D}$ with $9$ variables and ${\mathcal D}'$ with $10$ variables) 
and different group assignments. 
This led to change in the number of variables in Group~$9$ of Setting~$1$ and 
that in Group~$10$ of Setting~$2$ changed. 
For example, 
the anomaly score for Group~$9$ 
in the case of Setting~1~(see Table~\ref{tab03}) was estimated 
by substituting $t=\{z_{9}\}$ and $t'=\{z_{9},z_{10}\}$ to Eq.~(\ref{eqscr08}).
\begin{table}[htbp]
  \centering
  \caption{Left: Setting~1 of Section~\ref{experiment02}, 
change detection of constituent variables, 
where variable $z_{10}$ was newly generated in variable group~$9$. 
Right: Setting~2 of Section~\ref{experiment02}, 
where variable group~$10$ was newly generated.
}
    \begin{tabular}{rrr}
    \toprule
    Group & ${\mathcal D}$ & ${\mathcal D}'$ \\
    \midrule
    1     & $z_{1}$     & $z_{1}$ \\
    ${\vdots}$     & ${\vdots}$     & ${\vdots}$ \\
    8     & $z_{8}$     & $z_{8}$ \\
\midrule
    \multirow{2}[0]{*}{9} & $z_{9}$     & $z_{9}$ \\
          & None  & $z_{10}$ \\
    \bottomrule
    \end{tabular}%
\hspace{10mm}
    \begin{tabular}{rrr}
    \toprule
    Group & ${\mathcal D}$ & ${\mathcal D}'$ \\
    \midrule
    1     & $z_{1}$     & $z_{1}$ \\
    ${\vdots}$     & ${\vdots}$     & ${\vdots}$ \\
    9     & $z_{9}$     & $z_{9}$ \\
\midrule
    10     & None     & $z_{10}$ \\
    \bottomrule
    \end{tabular}%
  \label{tab03}%
\end{table}%
%
\par
As a kernel between variables, 
we used a covariance matrix. 
To compare a kernel matrix with an existing kernel between matrices, 
we used the Matrix Kernel and the Probability Product Kernel~(PPK)\cite{JKH04} 
as kernels between kernels~(i.e. kernels between matrices).
As mentioned in Section~\ref{introduction}, 
PPK can take 
different dimensional matrices as inputs.
We iterated the random data generation $100$ times to estimate the anomaly scores.

\subsubsection{{\bf Experimental Results}}
\par
%
\begin{figure}[htbp]
\vspace{-5mm}
\begin{center}
\includegraphics[width=89mm]{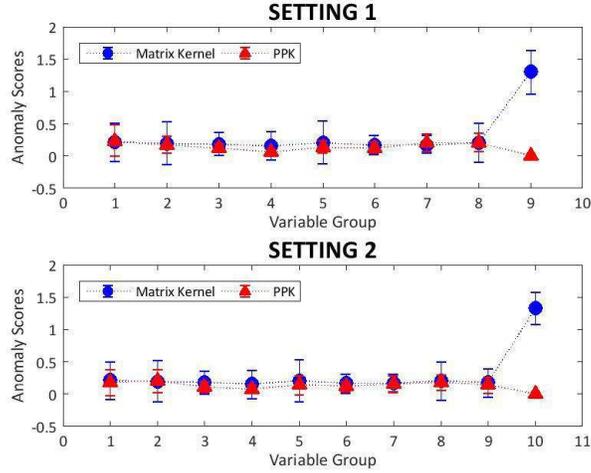}
 \caption{
Experimental results of Section~\ref{experiment02}, 
change detection of constituent variables.
Horizontal and vertical lines represent the 
variable group and the anomaly score of the variable group respectively.
A shown point and an error bar represent 
the mean and the standard deviation~(${\pm}1{\sigma}$) respectively. 
}
\label{fig03} 
\end{center}      
\end{figure}
The experimentally obtained results are presented in Figure~\ref{fig03}.
As the figures show, 
it is readily apparent that DKS is effective for 
systems in which the numbers of variables change 
for the following reasons.  
%
\par
As shown in the figure, we were able to estimate anomaly scores for datasets with different numbers of variables, using DKS.
DKS with the Matrix Kernel was able to detect the changes of assignments as anomalies successfully. 
The anomaly scores for Group~$9$ in Setting~$1$ and Group~$10$ in Setting~$2$ 
had sufficiently higher than those for the other groups. 
The results also indicate that 
the Matrix Kernel is more effective than the PPK. 
This is because 
DKS with the Matrix Kernel detected 
changes of the assignments as anomalies 
whereas DKS with the PPK failed to detect such changes.
\subsection{Anomaly Detection and Localization 
in Economic Time Series}\label{experiment03}
\subsubsection{{\bf Experimental Setting}}\label{exp03}
\par
DKS was applied for economic time series in this experiment. 
Namely, we conducted anomaly scoring for a single variable in the time series and the entire system simultaneously by using DKS.
%
\par
We used twenty economic time series consisting of 
nine FX~(foreign exchange) time series 
and eleven stock index time series as input. 
These FX time series include 
USDAUD, USDBRL, USDCAD, USDEUR, USDGBP, USDHKD, USDJPY, USDKRW, and USDRUB. 
The FX time series took values per 1 USD, whereas the stock index time series consist of AORD, BVSP, CAC40, DAX, DJI, FTSE100, Hang Seng, KOSPI composite, N225, RTSI, and TSX composite. 
The time series were observed once a day from 1st January 2004 to 31st December 2008.
\par
We decided to use the economic dataset for the following reasons. 
First, it includes the great depression starting in September 2008, which was triggered by the bankruptcy of Lehman Brothers. 
We suspected that DKS would detect this economic disorder as the anomaly. 
Second, it is expected that anomalies appear in response to changes in the relationships between some variables~(some time series),  as currencies and stocks strongly correlate with each other.  
\par
We applied DKS to the dataset with the following settings. 
We set time window width as $50$ days, where the $n$-th window consisted of the time series from $t\,=\,n-49~[day]$ to $t\,=\,n~[day]$. 
By comparing the $n$-th and the $(n-1)$-th windows, 
we estimated the anomaly scores of the system and the individual variables.
We designated the scores as ``scores at $t=n$''.
We used a correlation matrix and a Matrix Kernel respectively as a kernel between variables and that between matrices. 
\subsubsection{{\bf Experimental Results}}
\par
\begin{figure*}[hbtp]
\begin{center}
 \includegraphics[width=160mm,clip]{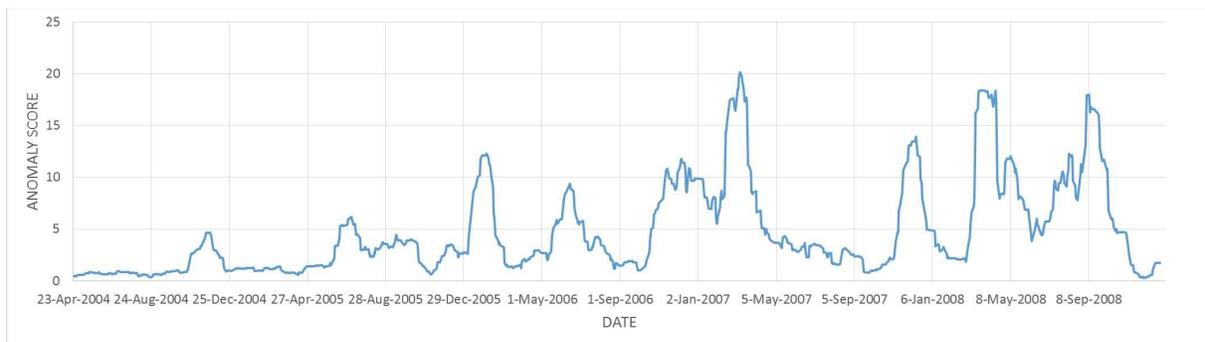}
\caption{Experimental results of Section~\ref{experiment03}, 
anomaly detection from the economic time series.
Anomaly score time series of the entire system. The horizontal and the vertical line represent time stamp and system anomaly score respectively.}
\label{A_fig3_1} 
\end{center}      
\end{figure*}
There were several peaks found in the system anomaly score time series~(Figure~\ref{A_fig3_1}). 
One of the peaks appeared on 8th September 2008, which was a week before Lehman Brothers announced its bankruptcy. 
The system anomaly score remained relatively high from November 2007 to October 2008 when the subprime mortgage crisis were ongoing from 2007. 
These results suggest that DKS successfully detected changes in the relationship among economic time series caused by the economic disorders as anomalies.
\par
\begin{figure}[tbp]
\begin{center}
 \includegraphics[height=67mm,clip]{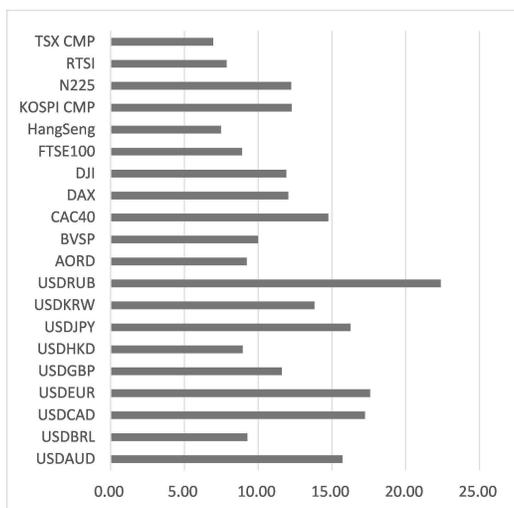}
\caption{Experimental results of Section~\ref{experiment03}, anomaly localization from the economic time series.
Anomaly scores of the individual time series on 8th September 2008, a week before the bankruptcy of Lehman Brothers. The horizontal line represents variable-wise anomaly score.}
\label{A_fig3_2} 
\end{center}      
\end{figure}
%
The variable-wise anomaly scores on 8th September 2008 are shown in Figure~\ref{A_fig3_2}.
The anomaly score of USDRUB was the highest among the variable-wise scores. 
USDRUB continued to decrease stably around the day of the bankruptcy while the other stocks and the currencies fluctuated significantly. 
This difference in trend could produce the high anomaly score of USDRUB.
\par
Here, we observed the behavior of DKS only qualitatively because we could not know what the anomalies to be detected were.  
However, the experimental results suggest that DKS is simultaneously applicable to change detection~(anomaly detection in a sense) by using the system anomaly score, and localization by using the variable-wise anomaly scores.

\section{Summary}\label{summary}
\par
We have developed the new anomaly detection method, Double Kernelized Scoring (DKS). 
This is a unified method to perform anomaly detection and localization simultaneously in a strongly correlating system with a changing number of elements. 
For comparing matrices with different dimensions, we have proposed a new kernel function, Matrix Kernel. 
The Matrix Kernel is defined between square matrices that might have different dimensions. 
We have demonstrated the effectiveness of DKS and Matrix Kernel through the experimental results using three datasets.

\section*{Acknowledgements}
We would like to thank Dr. Makoto Fukushima for fruitful discussions and comments on an earlier version of the manuscript. 


\end{document}